\documentclass[lettersize,journal]{IEEEtran}
\usepackage{amsmath,amsfonts}
\usepackage{algorithmic}
\usepackage{algorithm}
\usepackage{array}
\usepackage[caption=false,font=normalsize,labelfont=sf,textfont=sf]{subfig}
\usepackage{textcomp}
\usepackage{stfloats}
\usepackage{url}
\usepackage{verbatim}
\usepackage{graphicx}
\usepackage{cite}
\usepackage{authblk}

\begin{document}

\title{Cognitively Inspired Components for Social Conversational Agents}

\author{Alex Clay, Eduardo Alonso, and Esther Mondragón}

\affil{Artificial Intelligence Research Centre (CitAI) \protect\\ Department of Computer Science \protect\\ 
City, University of London}

    \markboth{\tiny This work has been submitted to the IEEE for possible publication. Copyright may be transferred without notice, after which this version may no longer be accessible.}%
{Cognitively Inspired Components for Social Conversational Agents}

\maketitle

\begin{abstract}
Current conversational agents (CA) have seen improvement in conversational quality in recent years due to the influence of large language models (LLMs) like GPT3. However, two key categories of problem remain. Firstly there are the unique technical problems resulting from the approach taken in creating the CA, such as scope with retrieval agents and the often nonsensical answers of former generative agents. Secondly, humans perceive CAs as social actors, and as a result expect the CA to adhere to social convention. Failure on the part of the CA in this respect can lead to a poor interaction and even the perception of threat by the user. As such, this paper presents a survey highlighting a potential solution to both categories of problem through the introduction of cognitively inspired additions to the CA. Through computational facsimiles of semantic and episodic memory, emotion, working memory, and the ability to learn, it is possible to address both the technical and social problems encountered by CAs.

\end{abstract}

\section{Introduction}

Conversational Agents (CA), are software systems intended to engage in natural language interactions with humans \cite{ahmad-2022}. Essentially, a CA communicates with a human user using utterances designed to be in line with what might be stated in human-human communication. Commonly seen CAs include software like chatbots, virtual personal assistants,  and more recently OpenAI's ChatGPT. 

The uses for CAs span a wide variety of industries including: commerce \cite{munnukka-2022}, entertainment \cite{zhou-2018}, mental health \cite{ahmad-2022, beredo-2022}, education especially in regards to language learning \cite{10.1007/978-3-030-90179-0_35} and more. This can be in the form of a shopping assistant that suggests products based on a user's interests, or a social interaction asking about the user's day and mood. CAs provide a unique way of interaction with humans, due in part to the adaptability of the concept.

CAs widely belong to one of two categories based on their purpose. A task oriented CA is tailored to a specific domain such as booking a flight. An example is Google's DialogFlow, which uses a specified flow designed to extract relevant information through the use of intents, variations, and pre-written responses \cite{sabharwal-2020}. Conversely, a non-task-oriented CA such as XiaoIce \cite{zhou-2018} is more common in ’chit-chat’\cite{yan-2018}. 

The components considered in this paper address two key categories of problems encountered by conversational agents: poor response quality and the perception of CAs as social actors. 

Widely speaking, CAs encounter two primary problems based on the technical approach followed in their construction, that of limited scope with retrieval agents \cite{yan-2018,beredo-2022} and until recently, the nonsensical answers of generative agents \cite{yan-2018, ma-2022}. With the introduction of more advanced CAs, nonsensical answers pose less of a problem as  generative agents are able to produce coherent responses, although they still suffer in situations where the interaction becomes more complex or drawn out \cite{ahmad-2022}. 

As a result of being perceived as social actors \cite{nass-1994}, CAs encounter the additional problem of contending with social expectations. When handled incorrectly, the result can be poor interaction quality or perceived threat on the part of the user. Contending with social expectation additionally requires the consideration of believability and anthropomorphization of the agent by the user, to avoid further pitfalls.

As such, this survey presents an overview of conversational agents with the perspective that a potential solution of both categories of problems is to augment a CA with cognitively inspired components. Such a solution would not only result in a generation process more akin to that of a human, but potentially yield more similar responses as well. This paper's contribution lies in the unique focus on a merger of concepts from cognitive science and their potential benefit to CAs.

Five cognitively inspired components are proposed for this purpose: semantic memory, episodic memory, emotional mimicry, working memory, and the ability to learn. Semantic and episodic memory provide contextual information and personal experience respectively, enabling a consistent and informed basis by which answers are produced. Emotional values would allow for more human-like answers while reducing problems caused by a lack of perceived social emotion on the part of the user. Additionally, when combined episodic memory and emotional components could provide means by which the user perceives the CA to possess empathy. Working memory provides context for the interaction at hand. Finally, the ability to learn will allow new information to be actively integrated, improving on what was given at training time and aiding in the perception of intelligence.

The rest of the survey will outline the aforementioned concepts and problems in more detail. Firstly, CAs are covered generally, followed by explanations of retrieval and generative based methods for CAs.  Next, the paper covers what it means for computers to be considered social actors, and the consequences of social expectations, anthropomorphism, and believability in regards to CAs. Finally, the concepts of semantic and episodic memory, emotion, working memory, and the ability to learn are addressed in regards to the concept itself as well as computational representations and their relation to CAs. Though the computational representations in regards to CAs are thus far independent, this survey takes the perspective that a combination would yield potentially more benefit than the sum of the components individually.

\section{Conversational Agents: The Technical Problem} 
A Conversational Agent (CA) is a program that processes and responds to a natural language input from a user. \cite{ allouch-2021}, pp. 3, in particular distinguishes a CA as "a dialogue system that can also understand and generate natural language content, using text, voice, or hand gestures, such as sign language." 

CAs are primarily approached in one of two ways: generative or retrieval based systems \cite{yan-2018, beredo-2022, ahmad-2022, ma-2022}, which are discussed in detail in later sections. A third form of CA, though less commonly encountered, is a rule-based system, which uses a pre-defined rule set to provide precise answers \cite{bilquise-2022}. Rule-based systems are not as prevalent as retrieval or generative CAs, and therefore are not covered in further depth in this paper.

Another qualifier often applied to CAs is that of being single-turn or multi-turn. Single turn CAs create responses based on the prior input, or the utterance that prompts the response, whereas multi-turn CAs produce responses based on multiple previous inputs, sometimes referred to as the dialogue context \cite{kim-2019}. 
Additionally, a conversational agent can be Disembodied (DCA) and simply be a text-chat or Embodied (ECA) with a visual representation, often designed to appear human and convey nonverbal signals \cite{perez-marin-2011}.

\subsection{Retrieval Conversational Agents}
A retrieval conversational agent is one that returns an answer for the given query from a bank of pre-written responses. Single and multi-turn matching models are two classifications of retrieval agents \cite{boussaha-2019}. Single turn models tend to feature an Long Short-Term Memory (LSTM), and do not explicitly differentiate the contexts. Conversely, in multi-turn models, the potential response is checked against each utterance in the context, and an aggregate score is produced for the response.

LSTMs, along with Gated Recurrent Units (GRU) \cite{cho-2014} are commonly seen forms of Recurrent Neural Networks (RNNs) \cite{hochreiter-1997}. RNNs are neural networks that employ recursion on sequential data. LSTMs resolve both the vanishing and exploding gradient problems of simple RNNs, where the gradients reach very small values and leave layers largely unchanged, and where gradients accumulate over backpropagation which result in large updates to the weights. The LSTM is able to resolve these problems by learning long-term dependencies, through four different layers: input gate, hidden state, forget gate, and output gates \cite{hochreiter-1997}. Conversely, a GRU only possesses a reset and update gate. Fewer parameters results in faster training than that of LSTMs \cite{cho-2014}.

Additionally, another means for retrieval systems is chain-based matching, in which the modelling of the second sentence is aware of the first, as compared with pairwise matching which compares sentences on a word-by-word basis. At the basis of retrieval systems is the problem of matching the input to a preexisting candidate response \cite{yan-2018}.

Retrieval based systems are popular amongst commercial CA solutions. For instance, Dialogflow is a system from Google which allows the creation of domain specific chatbots, which are able to fallback to use the Google knowledge base to find relevant documents. In the system, a flow is designed to extract relevant information through the use of intents, variations, and pre-written responses \cite{sabharwal-2020}. This system produces chatbots more in line with retrieval based systems, using utterances from the user to trigger domains pecific flows. However, this yields limited capability in non-service based tasks that would not follow a specified flow. This is similar to the Alexa system produced by Amazon AWS \cite{aws-no-date} that also uses intents. 

Outside of commercial solutions, research continues into retrieval CAs. Wu and colleagues \cite{WU2018251} proposed a topic-aware RNN, using attention to weigh the message and response in conjunction with their topic. This approach was found to place responses with richer content in higher rank. Another paper \cite{moore-2021} discerned that a self-supervised contrast model for response selection out preformed that of binary and multi-class classification. 

Where retrieval agents see a benefit in the nature of their individual responses being written by humans, they suffer due to issues of scope. Generative agents, conversely see the opposite.

\subsection{Generative Conversational Agents}

Generative CAs are those that produce a response on-demand to a given input. A prevalent means for their creation is through the use of an encoder-decoder framework \cite{zhang-2020} in which both the encoder and decoder are typically LSTMs or GRUs. 

Encoder-decoder refers to a model architecture that is frequently used in sequence-to-sequence (seq2seq) tasks. The encoder takes an input and uses it to create a context vector, a numerical representation of the input sequence. This context vector is then passed to the decoder along with a previously decoded word, the result from which is a token \cite{yan-2018}. In the case where the result is a decoded word, the word along with any preceding it are fed back into the decoder in order to generate the next word until a stopping token is reached. 

Mathematically this is represented as the following from Yan \cite{yan-2018} where \( (x_1, x_2, ..., x_n)\) are the embedding inputs, \((h_1, h_2, ... h_n) \) are the hidden representations, \(c_t\) is the context vector, \(y_{t-1} \) is an embedding of a previously decoded word, \( [c_t;y_{t-1}] \) is the concatenation of the previous two symbols, \(o_t\) is the output probability distribution, and \(s_t\) is the decoder state vector: 

\begin{equation}
    Encoding: h_t = Encoder(h_{t -1}, x_t)
\end{equation}
\begin{equation} 
Decoding: s_t = Decoder(s_{t-1}, [c_t; y_{t-1}])  
\end{equation}
\begin{equation}
y_t \sim o_t = p(y_t | y_1, y_2, . . . , y_{t - 1}, c_t) = softmax(W_o*s_t)  
\end{equation}

A seq2seq task is one where many inputs are mapped to many outputs, or given an input sentence, produce an output sentence \cite{sutskever-2014}.

However the encoder-decoder framework is rarely where the development concludes. Feature-based and fine-tuning were noted as two of the primary strategies for the application of pre-trained language representations \cite{devlin-2019}. Feature-based approaches use pre-trained representations in addition to task-specific architectures. Conversely, the fine-tuning approach minimises task specificity and tunes all pre-trained parameters instead. These language models are typically unidirectional, which poses problems in sentence level tasks as they are unable to incorporate bidirectional context.

\begin{figure}[!t]
\centering
\includegraphics[width=2.5in]{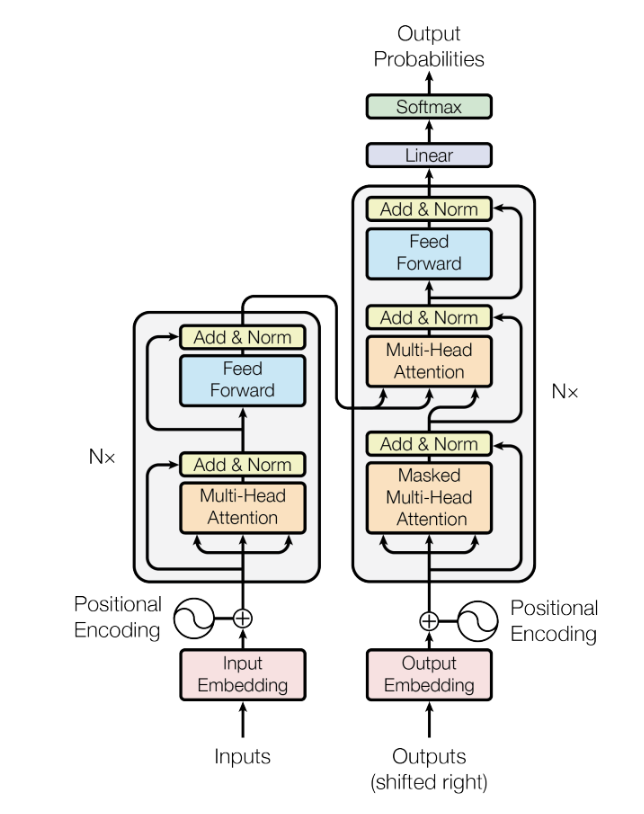}
\caption{Transformer architecture, which can possess any number of encoders and decoders, the original paper used 6 of each \cite{vaswani-2017}.}
\label{fig_trans}
\end{figure}

In order to understand models like BERT, is is first necessary to introduce transformers. The transformer \cite{vaswani-2017}, though similar to an RNN, removes the processing of the data in order, which allows for processing to happen concurrently and reduce training time. In the original publication, the model employed an encoder and decoder stack as outlined in Fig. \ref{fig_trans}. 

As the embeddings are not taken sequentially, a positional encoding becomes necessary in order to store the ordering. An important facet of transformer models is that of attention. Attention allows for other tokens to influence how the current token is interpreted. In a response generation task for example, attention would allow for both the word before and after the current word to help define its encoding. Self-attention, creates an attention-based vector of each word in the input in order to handle the contextual relationships between the words in the input. Mathematically this is represented by the following, where  the input contains queries and keys of dimension \(d_k\), a matrix \(Q\) of queries,  matrices \(K\) and \(V\) of keys and values respectively. :
\begin{equation}
    Attention(Q, K, V) = softmax(\frac{QK^T}{\sqrt{d_k}})V
\end{equation}

Multi-head attention is also utilised in transformers, which creates multiple attention-based vectors for each word in the input. The outlined architecture is that of the paper \cite{vaswani-2017}, however not all transformers are constructed with the same architecture. 

\begin{figure}[!t]
\centering
\includegraphics[width=2.5in]{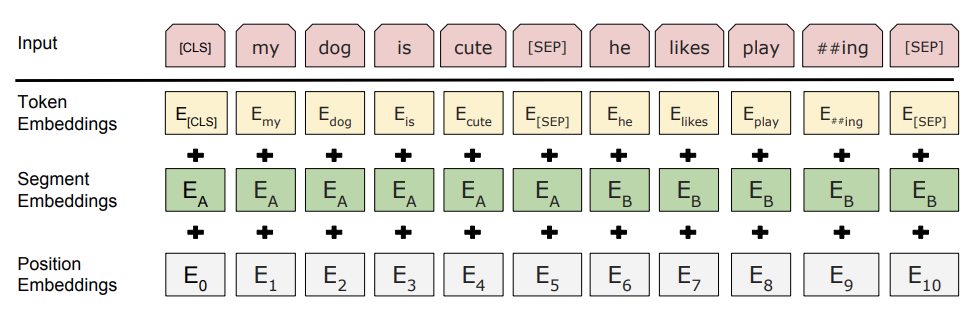}
\caption{BERT Token Embeddings, depending on the pre-training of the embeddings, such models can be used to translate between languages or generate a response to a given input \cite{devlin-2019}.}
\label{fig_token}
\end{figure}

Bidirectional Encoder Representations from Transformers (BERT), was the first of a series of models based on the principle of using unlabelled text to train deep bidirectional representations, with joint conditioning on left and right context on all layers. This approach achieved state-of-the-art performance on both sentence and token level tasks at the time of its release. Distinctly, BERT possesses a unified architecture, and was pre-trained using masked language model and next sentence prediction, the tokenisation for which can be seen in Fig. \ref{fig_token}. Moreover, the pre-training data focused on a document-level in order to use longer sequences. Consequently, BERT was able to cope with a wide variety of natural language processing tasks using the same model \cite{devlin-2019}.

\begin{figure}[!t]
\centering
\includegraphics[width=2.5in]{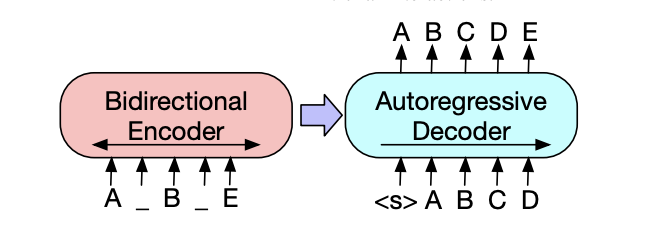}
\caption{BART framework: BART contains both Bidirectional Encoder, characteristic of BERT, and an auto-regressive decoder, characteristic of GPT \cite{lewis-2019}.}
\label{fig_bert}
\end{figure}

A BERT-VHRED model \cite{zhao-2019}, pre-trained with the purpose of conversational generation, was found to preform well in regards to perplexity, which is essentially the uncertainty or text fluency of the model, while avoiding the problem of overly generic responses present from the standard BERT model. 
BART \cite{lewis-2019} is a denoising autoencoder for pre-training sequence-to-sequence models, which can be seen as a generalisation of BERT. BART is especially noted for effectiveness in text generation. BART contains both Bidirectional Encoder and an auto-regressive decoder which uses information from previous timesteps to generate the current one, as illustrated in Fig. \ref{fig_bert}. 
Though neither BERT nor BART are directly intended for use in CAs, they provide a crucial grounding for the understanding of generative models. 

ChatGPT, a release from OpenAI, is a high quality chat generation system, developed using reinforcement learning atop the InstructGPT model \cite{ouyang-2022}. Though possessing no formal publication at the time of this paper’s writing, the informational page on the system outlined that one of the primary shortcomings was answers that sound plausible but are ultimately incorrect or nonsensical. Additionally, a slight change in phrasing can be the difference between the model being able or unable to produce a correct answer. Both of these issues are present when comparing retrieval and generative systems in different forms. InstructGPT, the underlying basis for ChatGPT, was designed with the intent of creating a model aligned with the user intent. This was done through human feedback on a number of tasks. The model was trained using demonstration data written by human labelers to provide an example of ideal behaviour, the output from this model was then labeled based on the preferred output for a given input which in turn trained a model to anticipate the preferred output. This second model’s output was then used to provide a scalar reward, optimised with the PPO algorithm \cite{ouyang-2022}. 

In the last year, OpenAI released GPT-4 \cite{openai-2023B}. Though able to meet human-level performance on evaluations like the bar exam for law, GPT-4 experiences a similar set of problems to that of prior GPT models, namely: limited context, information hallucination, and does not learn from experience. However one of the primary aims of GPT-4 was to have results that behaved predictably across multiple metrics which was achieved along with better adherence to desired behavior. 
When viewed as a whole, generative CAs possess a nearly reflected set of benefits and pitfalls when compared to that of a retrieval system. Generative systems benefit by avoiding limitation in scope due to a pre-written bank of responses. However, the quality of response is occasionally classed as lower than retrieval CAs, due in part to the responses being too general, nonsensical, or lacking in fluency \cite{yan-2018,beredo-2022,ahmad-2022, ma-2022}.

Despite the popularity of BERT and similar models, other networks continue to be investigated in regards to CAs. \cite{kim-2019} used Generative Adversarial 
Networks (GAN) based on Dual Wasserstein GANs to produce a multi-turn CA, by using one GAN for dialogue modelling and another for response generation. The results noted a higher quality of response than previous state-of-the-art models, but a lack of consistent personality due to the wide variety of data sources \cite{kim-2019}. Similarly, Convolutional Variational Auto-Encoders (CVAE) have been investigated as a means for yielding more diverse responses, yet the model is noted for struggling to do so without augmentation \cite{gao-2019}. 

Conversely to retrieval CAs, generative CAs are able to handle input regardless of scope, but may encounter issues with nonsensical or poor quality answers. As a result, both retrieval and generative CAs encounter problems when needing to meet the user's social expectations. 

\section{Computers As Social Actors: The Social Problem}

As early as 1994 \cite{nass-1994}, research has established that computers are perceived as social actors, and as such, are expected to adhere to social rules. Generally speaking, a social actor is one that acts based on understood rules of interaction, or social expectations. It has been shown that humans will apply these social expectations to computers as a natural response to social situations \cite{nass-1994}. The extension of social expectations to computers and subsequently CAs poses a unique problem as the failure to adhere to such expectations or display social emotion can lead to the user feeling uncomfortable, threatened, distrustful, and withholding information \cite{zhang-2021, pelau-2021}.

Moreover, it has been noted that CAs should possess social characteristics to avoid frustration and dissatisfaction in interaction \cite{chaves-2021} and additionally, in cases where a model might otherwise interact successfully, it may still fail due to poor social abilities \cite{see-2021}. A similar impact has been shown to exist in communication between humans where it has been seen to be crucial to adhere to social expectations in interaction. Briot et al. \cite{briot-2020}, pp. 1, discerned that, in reference to people with Autism Spectrum Disorder, “...a lack of social skills may lead to negative reactions from others, which promote negative beliefs and avoidant behaviors in social situations in the general population”. Given that avoidant behaviour is often observed when another human does not adhere closely enough to the societal expectation of a given interaction, it is unsurprising to see a similar outcome when non-human interactants do not comply with social expectations. 

The Computers as Social Actors theory, which states that it is expected that AI will adhere to the rules of society \cite{pelau-2021} further supports the notion that CAs must adhere to social expectation. Therefore, a failure to behave along expected social lines would lead to the same negative reactions that are displayed to atypical human behaviour. Moreover, it has been determined that consumers interact for a longer time with a more humanoid AI \cite{pelau-2021}. Additionally, in communication with agents, humans may withhold information and be distrustful if the environment lacks social emotion, something which can be mitigated by meeting the user’s social needs \cite{zhang-2021}. As such, in order for an agent to have a positive interaction with the user, it must adhere to social convention. 

Social characteristics also bring about social biases. It is expected that a CA communicating socially would be subjected to social bias as a person would \cite{tay-2014}. These social biases in turn result in biased associations regarding competence, authority, trustworthiness, and other socially relevant features. These biases are not entirely without potential benefit, as they impact how much an interaction is seen as easy, pleasant, engaging, and effective, and ignoring them could result in poor quality interactions\cite{fossa-2022}.

When considering what is important in social communication with CAs, there are many potential routes. Van Pinxteren and collaborators \cite{van-pinxteren-2020} presented a review of articles to determine what human-like communicative behaviours improve relational results. The findings were that two general areas became apparent: modality, which covers verbal and non-verbal behaviors and appearance, as well as footing which indicates similarity and responsiveness. Additionally, Rheu et al. \cite{rheu-2021} found that five factors impacted trust of a CA: social intelligence of the agent, voice characteristics and communication style, look of the agent, non-verbal communication, and performance quality. These factors and their influence were also shifted depending on the user’s demographic. 

It is also worth noting that people alter their language when they are knowingly interacting with AI. When comparing instant messaging interactions with those of a human and Cleverbot, people sent a higher number of shorter messages with the AI in addition to using a smaller vocabulary and more profanity. The study \cite{hill-2015} concluded in finding that language skills are transferable regardless of who the human is interacting with, and importantly that the increase in messages with the chatbots disproved the notion that people were less confident or comfortable with the chatbot, but rather adapting their behaviour to match that of the chatbot \cite{hill-2015}. This could imply that, though CAs are expected to adhere to social convention as social actors, that some allowances are made when it is acknowledged that the other party is non-human. 

Unlike with human to human interaction, when interacting with CAs, regardless of perceived social capability there is still the problem of handling the user's anthropomorphism of the CA and subsequently believability. 

\subsection{Anthropomorphism}

Anthropomorphism is can be defined as the projection of human characteristics, be it physical or behavioural, onto something non-human, and is a common behaviour among humans \cite{skjuve-2019, ruane-2019}. \cite{araujo-2018} While the application of human characteristics to an agent would increase its believability, there are also drawbacks. For instance, when an anthropomorphised agent makes a low-level mistake, it will make the user uncomfortable and fall into the uncanny valley \cite{zhang-2021}. Additionally, if the user anthropomorphizes a CA, then the absence of perceived empathy can result in the AI being seen as a threat \cite{pelau-2021}.

Perceived anthropomorphism, social presence, dialogue length, and attitudes are all factors which increase trust in Virtual Service Assistant (VSA) recommendations. With these, the VSA is able to be considered a social actor with a \textit{human touch} \cite{munnukka-2022}. A human-like agent, with human-like language or name, was found to have both higher mindless (unconscious evaluation) and mindful (conscious evaluation) anthropomorphism than that of disembodied conversational agents. However a humanlike vs a machinelike agent was not seen to have significant differences in social presence \cite{araujo-2018}. 

Sandry \cite{sandry-2015} argues in favor of 'tempered anthropomorphism’ in order to retain a 'sense of the otherness of the machine’. The paper explains how in a military setting, Explosive Ordinance Disposal robots are perceived as individuals despite clearly being non-human. It should, however, be noted that these robots are remotely human-controlled and therefore are not autonomous. However, unplanned behaviours, due to machine errors are read by teammates as communications in some cases. Therefore, in arguing to temper anthropomorphism, the author is more accurately referring to the acknowledgement of the agent as non-human, as they argue that machine-like robots would be able to provide forms of social communication instead of acting in place of a human or animal. 

Moreover, the uncanny valley theory, which is the cause of many concerns regarding anthropomorphism is facing scrutiny as to whether it is still applicable due to changing perceptions and improving technology. Mishra et al. \cite{mishra-2022} determined in their study that overall, the most anthropomorphic agents were also the most ‘well-liked’. Though the most ‘likeable’ agent also elicited the most sadness and second most fear, the study attributed this to the desire to impress the agent. This leaves the question as to whether the expectations for interaction with agents has changed or if the quality of humanness is high enough to avoid the uncanny valley entirely. Moreover, Jang discovered that in the case of AI influencers (social media influencers who are AI, rather than humans who talk about AI), eeriness was induced by the realism of their appearance but this did not reduce their likeability \cite{jang-2022}. The study also showed that overall attitudes were more favourable the more humanness the AI influencer displayed, a contradictory stance to that of the uncanny valley. Therefore, the commonly attributed notion that the more human-like an agent is, the more likely it is to fall into the uncanny valley is not necessarily accurate in regards to modern agents and their visual representations. Conversely, a recent study of a text chatbot compared with an avatar reading the chatbot text determined with the text only chatbot \cite{ciechanowski-2019} there was less negative affect and uncanny valley effect. As such, the application of the uncanny valley to CAs and even the continued use of the theory is somewhat inconsistent. 

That is not to say that anthropomorphism poses no risk. Without offsetting the problem of quality and scope in current CAs, perceived anthropomorphism would be detrimental rather than beneficial. Primarily resulting in reduced trust, as a lack of perceived empathy in an anthropomorphised agent can lead to the user feeling threatened \cite{pelau-2021}. Additionally, if an anthropomorphised agent were to make a low level mistake, such as a nonsensical utterance, it would cause the user discomfort \cite{zhang-2021}. One proposed solution is to reduce the anthropomorphism of the system \cite{sandry-2015}. However, reduced anthropomorphism is detrimental to the goal of accommodating a user’s application of social rules to the system, especially as CAs tend to be anthropomorphised \cite{ruane-2019}. Fundamentally, a less human-like conversational partner would struggle to maintain a coherent conversation, and a more capably maintained conversation would lead to the human user naturally anthropomorphising the CA \cite{skjuve-2019}. Therefore seeking to reduce anthropomorphism is not a particularly viable solution to the problems that can result due to anthropomorphic failure. 

One potentially key metric for evaluating anthropomorphism in CAs is the Godspeed questionnaire. Five key concepts are able to be evaluated: anthropomorphism, animacy, likeability, perceived intelligence, and perceived safety. The questionnaire was developed with the intention of providing means for the comparison of implementations within the field \cite{bartneck-2009}.  In 2015, a meta-analysis of the questionnaire found that it was one of the most used questionnaires in the human-robot interaction field \cite{weiss-2015}. While largely used in the frame of robot interaction, there is precedence for using the questionnaire to evaluate CAs \cite{liu-2022}.

\subsection{Believability}

Anthropomorphism and believability  are closely related. As believability is largely interpreted as the expectation of interacting with a human, and anthropomorphism is the projection of human characteristics onto something that is not human.  As such, it may hold that while it is possible to have anthropomorphism without believability, the reverse is not true. 

Both anthropomorphism and believability have a resounding impact on perception and behaviour when interacting with a CA. The Turing Test \cite{turing-1950}, as it is now known, is likely the earliest introduction of the concept of believability. The imitation game, at its simplest, seeks to know if a machine could imitate a person, and convince an ”interrogator” that it is the human, rather than the other player that actually is. As such, believability is largely categorised as whether or not a user believes they are interacting with a human rather than a program \cite{livingstone-2004}.

As early as 2000 it became clear that creating a believable agent required many facets, largely learning, emotions, and cognition \cite{el-nasr-2000}.  Rousseau and Hayes-Roth’s \cite{rousseau-1997} Cybercafe sought to determine how to make personality rich characters, and to understand how people would interact with such agents. The study showed that humans felt consistent characters to be believable, however this left the question of how to create such agents without pre-coded scenarios \cite{rousseau-1997}. However, Logacheva and colleagues \cite{logacheva-2020} discerned that a system could maintain a persona and consistency and still fail to converse well, highlighting the need for other additional components. Two aspects particularly noted for a believable agent in this respect, were that of emotions and social capability \cite{reilly-1996}. 

There has also been a significant correlation between believability of an agent with emotional variables as compared with one without \cite{sutoyo-2019}. This is furthered by El-Naser and collaborators \cite{el-nasr-2000} where it was determined that perceived emotion may play a crucial role in increasing the believability of an agent \cite{el-nasr-2000}. In order to simulate a personality, learning and emotional mimicry amongst others may be required. Moreover, the ability to learn had the most impact in terms of the agent seeming believable and intelligent. 

In a more recent study an AI with a physical form used a series of rules to discern what factors were important in solving its task. Through exposure to multiple possibilities, it used the rules to better understand its objective. Learning cumulatively from each version of the task, the robot was able to determine what factors were important on its own, therefore seemingly learning from its experience \cite{bhat-2020}. If an agent were to do something similar in interactions rather than a physical task, it poses the possibility of increasing its believability and allowing the agent to evolve. Moreover, PourMohammadBagher and collaborators \cite{pourmohammadbagher-2009} found that series of personality updates would allow for the model to evolve. Clear display of learning capability has also been seen to be positive in of itself. In a gameplay setting, it was determined that in-game training utilising the player saw results that were “easily recognizable, and sometimes delightfully surprising” \cite{cimolino-2019}, pp. 1. 

In order to fully address the social problem of user's social expectations, it is crucial to acknowledge the role of both anthropomorphism and believability. While it is not necessary for an agent to be believable to carry out a successful interaction, it is highly likely that the user will anthropomorphise the agent, at which point the agent must either be able to be believable or able to handle the potential results of anthropomorphism by providing social emotion and perceived empathy. We theorise that by adding cognitively inspired components such as episodic memory, emotion, and the ability to learn, it will be possible to benefit from the user's anthropomorphism of the agent without the problems that can result from it. 

\section{Cognitively Inspired Additions: A Proposed Solution}
Though the problems of the technical approach to CAs and social expectations of the users are unique, they share the commonality of both ultimately resulting in a poor interaction. In order to resolve such problems, research was conducted into what results in a good interaction, and the underlying cognitive concepts therein.

In recent years, there has been increasing interest in systems which derive inspiration from cognitive architectures. In their 2020 paper, Samsonovich \cite{SAMSONOVICH202057} outlined three key challenges in Biologically Inspired Cognitive Architectures (BICA). Firstly, the integration of “human-level artificial emotional intelligence” by which agents would use emotions in their decisions much like humans. Secondly, the ability to understand the context of what is happening. And finally, the implementation of human-like active learning, through the agent’s development of a system of values that will motivate its learning \cite{SAMSONOVICH202057}. 

It is likely no coincidence that such additions would also help to resolve both types of problems encountered by CAs. A potential means by which to address each of these challenges and problems is in  the form of episodic memory, emotional mimicry, working memory, and the ability to learn. 

\subsection{Semantic Memory}
Semantic memory is one’s knowledge of the world, holding concepts, facts, and beliefs \cite{yee-2014}. Semantic memory has been computationally represented in a number of different ways, each with a unique way of storing the information. One such method is through graphical representations. 

Knowledge Graphs (KG) are a means for representing knowledge through nodes and edges \cite{ji-2021}. 
In a KG, knowledge is represented through triples in the form of (head, relation, tail) where both the head and tail are nodes, and the relation is an edge. It is possible to use 'knowledge graph' and 'knowledge base' interchangeably. KGs have also attracted attention in recent years as a way to more closely mimic the structure of human knowledge. Heterogenous graphs especially,  possess different types of edges and nodes which allows for an additional degree of complexity, more similar to that of realistic semantic relationships \cite{zhu-2019}. KGs are able to extend beyond static information through the use of embeddings.  Knowlegde Graph Embeddings (KGE) present a means for retaining the semantic meaning of a triple in a low-dimensional vector mapping \cite{ji-2021}. One such means for embedding is that of TransE \cite{10.5555/2999792.2999923}. The principal behind TransE is that the head node should be embedded near the tail. 

KGs are not the only means for holding semantic information. A Bayesian network, for example, is a directed acyclic graph with the addition of a conditional probability distribution \cite{cheng-1997}. The Markov condition states that in a Bayesian network, every node is conditionally independent of those that are not parent or descendant nodes \cite{scanagatta-2019} and as such, densely connected networks contain less information about independence \cite{cheng-1997}. Extended treeaugmented naive classifier (ETAN), an extension of tree-augmented naive classifier (TAN) and Naive Bayes, at the time of the survey’s publication, was the most updated means of determining where a new observation should be stored, rectifying the problem encountered by Naive Bayes which often incorrectly estimated this \cite{scanagatta-2019}.

It is not uncommon to see the integration of semantic information into CAs. Grassi et al. \cite{grassi-2022} proposed a system for knowledge grounded dialogue flow based on an ontology for potential conversation topics, using a management system to determine the best topic to select based on the user’s input. When comparing keyword with keyword-category based matching systems, despite being more computationally taxing, the later provided more consistency and reduced the likelihood of random selection should the keyword not be found. .This was supported by the results of the study, finding that the keyword-category system was statistically more coherent than random selection and Replika a commercial chatbot. This study’s system was built with a manually encoded ontology for the keyword matching, and using the Cloud Natural Language API from Google for its Content Classification tool. A knowledge grounded conversational system, this study relies on internal and external information, that which the system already knows and what information it acquires during run time. The principal behind this is it would allow for more specific and natural dialogue based on knowledge about what the user’s input is referring to \cite{grassi-2022}. The study concluded that these systems were successful in avoiding dialogue that is only reactive.

While semantic memory, or the handling of facts has well been established, it is only more recently that another form of memory has been investigated. In 2015, it was found that through deep Q-learning, previous experiences could be generalised and used in comparison with novel experiences, a mechanism similar to that of animals \cite{mnih-2015}. This use of previous experience is what is known as episodic memory in cognitive science. 

\subsection{Episodic Memory}
The concept of episodic memory was first proposed by Tulving in the 1970s \cite{tulving1972episodic}, and was refined for his 2002 paper \cite{tulving-2002}. Crucially, episodic memory provides the ‘what’, ‘when’, and ‘where’ of events as they were experienced. This concept is also closely tied to that of ‘autonoetic’ and ‘noetic’ systems, or remembering and knowing respectively \cite{griffiths-1999}. Autonoetic is associated with episodic memory, and noetic with semantic. While Tulving did not provide a definition on the particular boundaries of episodes, Ezzyat and Davachi \cite{ezzyat-2011} discerned that events in episodic memory are usually delineated by spatial or temporal shifts. More recently evidence has suggested that episodic memory may be organised on narrative coherence \cite{cohn-sheehy-2021}.

Episodic memory in humans is crucial to decision making, as past experiences are used to inform novel situations. As detailed in Bar \cite{bar-2009}, some believe that behaviour is driven by experiences in memory, a notion which is also known as Bayesian Analysis. The paper also emphasized the importance of retaining information about the context in which something was learned, which would therefore aid in accessing the most relevant associations. Shohamy and Daw \cite{shohamy-2015} furthered this idea by showing that memories are often drawn on to generalise for use in the current task. The addition of episodic memory has also been determined to decrease the necessary episodes to determine a beneficial policy \cite{murphy-2020}. 

Murphy et al. \cite{murphy-2020} outlines three crucial properties for an effective memory system, in short to store a memory, be able to recall a memory from a partial cue, and to store and recall large quantities of experience. In order to do this, it was also highlighted that the feature vector space should be grouped by similarity, both in the nodes and their edges. Moreover, they outline a complete memory as possessing “the objects, entities, and relationship present at the time of encoding” \cite{murphy-2020}, pp. 2, all of which would be available from a partial cue. 

Some research has been done into providing an episodic memory component as a means for augmenting dialogue. Sieber and Krenn’s \cite{sieber-2010} proposed system allowed for previous interactions to be commented on, however these previous interactions focused on fact-based information and communication regarding such questions rather than as an integration of personal experience. 

The addition of episodic memory would allow for a system to more closely emulate that of the human decision making process, as well as allowing for the integration of personal experience. Moreover, memory and experience impact other systems that influence decision making.

\subsection{Working Memory}
Working memory is the system that provides handling and manipulation of information within the human mind for the present task. Evolved from the concept of short term memory, in some theories it possesses three components: the central executive which acts as an attentional-controlling system, the visuospatial sketchpad which hands visual images, and the phonological loop which, abstractly speaking, handles language \cite{baddeley-1992}. 
Baddeley \cite{baddeley-2000}, pp. 1, proposes that a fourth component is present in working memory, that of an episodic buffer. A system which ”... comprises a limited capacity system that provides temporary storage of information held in a multimodal code, which is capable of binding information from the subsidiary systems, and from long-term memory, into a unitary episodic representation.”. Additionally, Young and Lewis \cite{young-1999} proposed that working memory could not be studied independently of long term memory and learning. As such, when determining the construction of a computational model of working memory, episodic memory and learning are important to consider. 
Some research has been done into simulating working memory, though in the aim of a direct translation between the brain structure and a computational representation \cite{reser-2022}. The relevance of this to CAs lies in the potential for reducing issues of context loss, as well as providing  a means by which to more easily integrate information from conversations into a main knowledge base.

\subsection{Emotion}
According to El-Naser and collaborators \cite{el-nasr-2000}, memory and experience greatly influence emotion. 
Like memory, emotions are important to human intelligence and are integral to the decision making process \cite{mejia-2016}. Emotionally intelligent systems would complement humans, in that, human decisions are driven in part by emotion, and when decisions often have to be made with little or incomplete knowledge, where there are not memories, or little time to respond, adding emotions to an agent could more closely mimic the human process \cite{brennan-2020}. Approaches to implementing simulated emotion in agents have been varied. Fuzzy logic has been employed in order to allow a mixture of emotions \cite{el-nasr-2000}, an important capability, as humans can experience a mixture of emotions at different intensities concurrently \cite{asghar-2020}. Conversely, Ashgar and colleagues’s \cite{asghar-2020} Affect Control Theory uses the three facets Evaluation (E), Potency (P) and Activity (A) and predicts how a human in a social situation will respond to emotional stimuli, and can be used to try and avoid conflict in human interaction. Another approach, Sentiment Look-Ahead provides a reward based on the improvement to the user sentiment. This paper \cite{shin-2020} gives multiple model and equation combinations, yielding two which are considered to be more semantically similar to human responses. 

A widely used means for simulated and human emotional measuring is through valence, arousal, and dominance (VAD) \cite{russell-2003, russell-1980}. These dimensions have been determined to be the most important in regard to word meaning: valence indicating positiveness/negativeness, arousal indicating active/passive, and dominance indicating dominant/submissive. All of which aid in sentiment and emotional analysis. The scores for which are given in (V,A,D) scaling from 0 the lowest to 1 the highest \cite{mohammad-2018}.

Simulating emotions in CAs has seen more interest in recent years\cite{bilquise-2022}. Lexicon based and machine-based are the two primary means of handling emotional mimicry, the first uses a dictionary to capture emotion and the second uses a classifier. The lexicon approach was most used in approaches that address accurate capture of the simulated emotion. Some methods noted in the survey \cite{bilquise-2022} have also used a combination of the two approaches. 

Srinivasan et al. \cite{srinivasan-2021} for instance, used embeddings with simulated emotion values through VAD scores in a bidirectional seq2seq model. Additionally the study proposed an additional internal reward dubbed Emotional Intelligence which was based on the source and generated text's affective dissonance. 

Peng et al. \cite{peng-2019} proposes a topic-enhanced emotional conversation generation model. Using a LDA model to obtain topic words of the input sequences, then using a dynamic emotional attention mechanism to gain relevant information from the input texts and topics. The resulting decoder model is able to generate responses related to a specified emotion. Zhou et al.’s \cite{zhou-2017} paper regarding an Emotional Chatting Machine also makes use of the encoder-decoder framework. The system is constructed with three novel mechanisms: embedding simulated emotion categories, change of implicit internal simulated emotional states, and explicit simulated emotional expression with an external simulated emotion vocabulary. The result was a system capable of contextually and emotionally appropriate responses. 

Also important in human communication is perceived empathy, of which experience and emotion are often considered key.  

\subsection{Empathy}

Empathy can be widely defined as the psychological capacity that allow humans to understand what another is thinking and feeling, and in turn to emotionally engage with one another. In some cases empathy is viewed as ”social glue” that allows the establishment of social relations \cite{stueber-2019}. 

Empathy has been found to be important in communication with humans. A lack of perceived empathy in a CA can result in a user feeling threatened \cite{pelau-2021} and empathy is widely understood as a means for allowing people to connect with one another \cite{riess-2017}. However, the components of empathy vary: emotion sharing, perspective taking, and compassion \cite{depow-2021}, the perception and sharing of experiences, emotions, and needs \cite{riess-2017}, or more vaguely, the emotional, cognitive, and behavioural \cite{moudatsou-2020}. Despite some differences in the definition, all of these representations contain a reference to emotion, emphasising it as a key part of empathy. 

Chen et al. \cite{chen-2021A} introduced simulated empathy as an internal driving factor in an agent’s decision making, and found that mimicking empathy has long term benefits for the agent by allowing for its development within a community. Simulated empathy has also been emphasised as a way to elicit prosocial behaviours of agents \cite{chen-2021A}, some research even concluding that empathy is critical for compassion, which combined can motivate a response \cite{riess-2017}. Moreover, a correlation has been shown between opportunities to empathise and pro-social behaviour in humans \cite{depow-2021}. 

XiaoIce \cite{zhou-2018}, as of 2020, is a CA designed to be empathetic and social with optimisation for long-term engagement. With an average of 23 conversation turns per session, XiaoIce had longer interactions than other CAs at the time. Both Intelligence Quotient and Emotional Quotient were factored into the design as it was noted, that social skills of a certain level are important for a social chatbot. 
XiaoIce’s conversation engine layer is comprised of a dialogue manager, an empathetic computing module, a ’Core Chat’ which handles general conversation, and dialogue skills for domain-specific conversing. Importantly, the empathetic computing module is designed to interpret the user’s input along with the user themselves through emotion, intent, opinion on the topic, their background, and general interests. XiaoIce has been shown to be very successful in communicating with users in a way which encourages them to return to the system to continue conversing, and maintaining a persona that can be described as affable. However, it is worth noting that XiaoIce’s empathy was measured in regards to ”If a person enjoys its companionship (via conversation), we can call the machine “empathetic”” \cite{zhou-2018}, pp. 3, which is contrary to empathy as defined earlier.

Another implementation with a focus on empathy is that of the Sentient Embodied Conversational Agents (SECAs) \cite{tellols-2020} the system aimed to improve believability, through what they defined as “sentient capabilities”, namely personality, needs, and empathy. The system also included a means for memory, to avoid repetition. The SECA system is comprised of a number of modules: Personality, Needs, Conversational and Knowledge, Memory, Empathy, and NLP Modules. The SECA paper further highlights the importance of empathy and memory in social agents.

\subsection{Learning from Experience}
BlenderBot3 (BB3) is a system from Meta/Facebook that was intended to learn from its interactions\cite{shuster-2022}. The BB3 is primarily a transformer model combined with a number of seq2seq modules. Despite being composed of modules they are not independent as a single transformer model executes the modules. The modules primarily focus on three sections: internet search, long-term memory, and response generation. The internet search itself is composed of three modules, determining whether to search, generating the query, and completing the search. This allows for the system to supplement the existing knowledge base of the system dynamically based on its interaction with the user. The BB3’s long term memory system is similar to that of its internet access component, including a decision to access, and the actual accessing of the memory, with the addition of generating the long term memory. Finally, the BB3 can generate a knowledge based response, extract a relevant entity, and access a personal fact in long term memory with which to generate the final dialogue response. 

CAs like BlenderBot3 are being created with the ability to continually learn from their interactions \cite{shuster-2022}. Xu et al. \cite{xu-2022} notes that learning after deployment can improve skills connected to the pre-training and fine tuning data, as well as new skills. One particular example of continual learning is an agent designed for never-ending language learning (NELL) \cite{carlson-2010}. After 67 days and 66 versions, 242,453 beliefs had been promoted. The rule learner, a learning algorithm that learns rules in order to infer new relations from relations already present in the knowledge base, ran every 10 versions. Interestingly, the SOAR architecture \cite{laird-1987} is cited as similar work, a structure possessing working memory. As such, working memory is an underlying structure that can support a CA's learning by processing incoming experiences and allowing for later integration into the knowledge base.

\section{Conclusion}

This paper presents a survey highlighting the benefits of a conversational agent augmented with cognitively inspired components. Firstly, these components would offset the problems posed technically by the limitations of common approaches to CAs, be that the scope of retrieval agents or the potential nonsensical nature of generative agents, and the problem of context that impedes both. This is accomplished through the presence of semantic and episodic memory which provides a consistent grounding for generated responses and working memory which ensures that the context and information gathered throughout the interaction are maintained. 

Secondly, the additions address the user's requirements for social emotion, perceived empathy, and reduce problems resulting from the user's anthropomorphisation of the agent. This could be accomplished through the emotional integration providing emotional context and framing to the answers, the combination of episodic memory and emotion could also provide a means for the perception of empathy. Finally, the integration of learning will increase the overall appearance of intelligence, which in turn could improve the agent's believability. 

The integration of such components provides benefits that extend beyond those presented in this paper: the means for a CA that would be able to more gracefully recall previous interactions with a particular user, integrate new information, and maintain a consistent persona amongst others. By providing an underlying structure comprised of more than simply an large language model, the CAs response would be able to extend beyond the next most statistically likely word and immediate context, to potentially a CA capable of social communication beyond the capability of a GPT model, while requiring less data to run, and minimising re-training. 

At a high level, computational representation of the above concepts into a conversational agent could be achieved through a generative model, knowledge base, and emotional state handling. 

The knowledge base would contain the semantic and episodic memory, along with emotional values. This would provide contextual information to the generative model to use when formulating a reply, allowing for the agent's persona to be maintained, as well as potentially information about the user being interacted with. The responses and user inputs would also feed back into the knowledge base to allow for the integration of new information.  The working memory would act as a sub-component to the main knowledge base, holding the information for the current session and processing the information for later integration. The emotional state handling would augment the generated response to ensure that the agent's emotion stays consistent with the conversation and avoids responding only to the emotion in each given input. 

Research is currently being conducted into an implementation by which to test the theory proposed in this paper.

\bibliographystyle{IEEEtran}
\bibliography{ref}

\end{document}